\documentclass[conference]{IEEEtran}
\usepackage{times}

\usepackage[numbers]{natbib}
\usepackage{multicol}
\usepackage[bookmarks=true]{hyperref}

\usepackage{mathtools}
\usepackage{amssymb}
\usepackage{amsmath}
\usepackage{dblfloatfix} 
\usepackage{comment}
\usepackage{hyperref}
\usepackage{graphicx}
\usepackage{subcaption}
\usepackage{caption}
\usepackage{xcolor}
\usepackage{siunitx}
\usepackage[section]{placeins}
\usepackage[ruled,vlined,linesnumbered]{algorithm2e}
\DeclareMathOperator*{\argmax}{arg\,max}

\newcommand{\norm}[1]{\left\lVert#1\right\rVert}
\newcommand{\ie}{{i.e.}}
\newcommand{\eg}{{e.g.}}
\newcommand{\realR}{\mathbb{R}}
\newcommand{\circlegroup}{\mathbb{T}}

\usepackage{xcolor}
\usepackage{fancyhdr}

\usepackage[bottom,flushmargin,hang,multiple]{footmisc}
\makeatletter
\renewcommand\footnoterule{
\kern-3\p@
\hrule\@width.4\columnwidth
\kern2.6\p@}
\makeatother

\begin{document}

\title{Multi-Fidelity Black-Box Optimization for Time-Optimal Quadrotor Maneuvers}



\author{\authorblockN{Gilhyun Ryou, Ezra Tal, Sertac Karaman}
\authorblockA{Massachusetts Institute of Technology, Cambridge, Massachusetts 02139\\
Email: \{ghryou, eatal, sertac\}@mit.edu}
}


%

\maketitle

\begin{abstract}

We consider the problem of generating a time-optimal quadrotor trajectory that attains a set of prescribed waypoints.
The problem is challenging since the optimal trajectory is located on the boundary of the set of dynamically feasible trajectories.
This boundary is hard to model as it involves limitations of the entire system, including hardware and software, in agile high-speed flight.
In this work, we propose a multi-fidelity Bayesian optimization framework that models the feasibility constraints based on analytical approximation, numerical simulation, and real-world flight experiments.
By combining evaluations at different fidelities, trajectory time is optimized while the number of costly flight experiments is kept to a minimum.
The algorithm is thoroughly evaluated in both simulation and real-world flight experiments at speeds up to 11~m/s.
Resulting trajectories were found to be significantly faster than those obtained through minimum-snap trajectory planning.
\thispagestyle{fancy}
\let\thefootnote\relax\footnote{This work was partly funded by the ONR Grant N000141712670.}
\end{abstract}

\IEEEpeerreviewmaketitle

\section*{Supplementary Material}
A video of the experiments is available at \url{https://youtu.be/igwULi_H1Kg}.

\section{Introduction}



In recent years, fast navigation of autonomous vehicles has received increasing interest.
For several years an autonomous drone racing competition has been organized at IROS \cite{irosracingcompetetion}, and last year the AlphaPilot challenge introduced a similar competition to a more general audience~\cite{guerra2019flightgoggles}.
Currently and in the near future, state-of-the-art algorithms in estimation and control are reaching a level of maturity, such that the bounds of what is physically possible with a given vehicle are approached.
This presents the need for trajectory planning algorithms that fully exploit the capabilities of the vehicle and take into account the intricate limitations of vehicle dynamics, instead of relying on simplified models.

\begin{figure}
  \centering
  \includegraphics[width=0.48\textwidth]{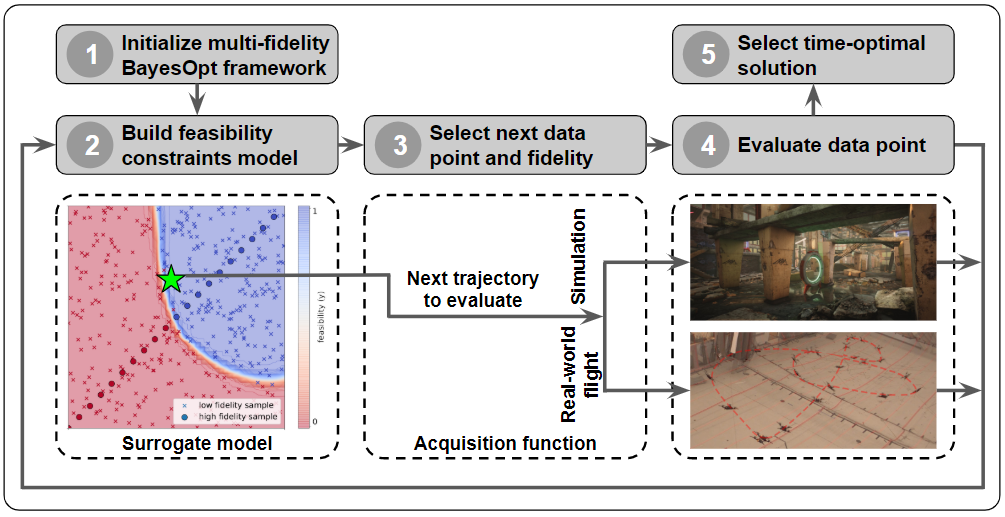}
  \caption{Overview of the proposed algorithm that models dynamic feasibility constraints based on simulation and flight data to efficiently find the time-optimal trajectory.}
  \label{fig:intro_diagram}
  \vspace{-5mm}
\end{figure}

In this paper, we consider the problem of generating, \ie, planning, a dynamically feasible, time-optimal quadrotor trajectory that passes through a given set of waypoints.
By definition, such an optimal trajectory is found at the boundary of the set of feasible trajectories.
Hence, precise knowledge of the dynamic feasibility constraints is required to find the time optimum.
This complicates the problem, as these feasibility constraints can become highly complex in light of high-acceleration flight and aggressive attitude changes, as required to achieve time optimality.
The demanding maneuvers affect flight dynamics, but also hardware and software for control and state estimation.
The resulting set of non-convex feasibility constraints with memory (i.e., depending on present and past states and control inputs)
cannot readily be incorporated in a typical trajectory planner for two main reasons.
Firstly, the feasibility constraints are not easily expressed in a convenient way, e.g., as constraints on an admissible set of control inputs and states.
Instead, the feasibility of the trajectory must be considered in a holistic manner.
Secondly, in most scenarios, precise modeling of these constraints is only possible through real-world experiments.
As these experiments aim to seek the boundary of the feasible set, they are risky and potentially costly and should thus be kept to a minimum.
Contrarily to this objective, many trajectory optimization schemes rely on a large number of evaluations.
These two issues form the main motivation for our algorithm, which uses a multi-fidelity optimization technique that can approximate the system feasibility constraints based on a limited number of experiments.
It uses a Gaussian process black-box model to classify candidate trajectories as feasible or infeasible and is thereby able to plan increasingly fast trajectories as the model improves. An overview of the algorithm is shown in Fig. \ref{fig:intro_diagram}.

This paper contains several contributions.
Firstly, we propose an algorithm for modeling of quadrotor feasibility constraints and generation of time-optimal trajectories based on Gaussian process classification.
Secondly, we extend the applicability of the multi-fidelity deep Gaussian process kernel from the regression problem to the classification problem in order to obtain a multi-fidelity Gaussian process classification algorithm that can incorporate evaluations from analytical approximation, numerical simulation, and real-world flight experiments.
Thirdly, we design an acquisition process specifically tailored to experimental robotics.
The acquisition function takes into account the additional cost of infeasible evaluations, as these may pose a threat to the vehicle and its surroundings.
Candidate data points are generated using minimum-snap perturbations in order to maintain feasibility.
Fourthly, we present an extensive evaluation of the proposed algorithm in both simulation and real-life flight experiments at speeds up to 11 m/s.
A video of these experiments accompanies this paper.
It is found that optimized trajectories are significantly faster than those obtained through minimum-snap trajectory planning.

The outline of the paper is as follows. In Section \ref{sections:preliminaries}, we present the problem definition, and preliminaries on trajectory planning and Bayesian optimization.
In Section \ref{sections:algorithm}, we detail our algorithm for the generation of time-optimal trajectories using iterative experiments.
In Section \ref{sections:experiment}, we present experimental results from both simulation and real-world flights.

\section{Preliminaries} \label{sections:preliminaries}



\subsection{Quadrotor Trajectory Planning}
In this work, we consider the problem of planning a time-optimal quadrotor trajectory that passes through $m+1$ given waypoints $\tilde {\mathbf{ p}} =  \begin{bmatrix}\tilde p^0 &\tilde p^1& \cdots& \tilde p^m\end{bmatrix}$.
Each waypoint consists of a prescribed position and yaw angle, \ie, $\tilde p^i = \begin{bmatrix}
{{}\tilde p^i_r}^T & \tilde p^i_\psi
\end{bmatrix}^T \in \realR^3 \times \circlegroup$, where $\circlegroup$ indicates the circle group.
A trajectory $p : \realR_{\geq 0} \rightarrow \realR^3 \times \circlegroup$ is a continuous function that maps time to position and yaw, \ie, $p(t) = \begin{bmatrix}
{{}p_r(t)}^T &  p_\psi(t)
\end{bmatrix}^T$.
It is constrained to start at $\tilde p^0$ and subsequently attain the remaining waypoints in order, so that we can define the following time-optimal planning problem:
\begin{equation}
\begin{aligned}
&\underset{p,\; T,\; \mathbf{x}\in \realR^m_{\geq 0}}{\text{minimize}}
 & & T\\
  &  \text{subject to} & & p(0) = \tilde{p}^0, \\
&   & & p\left(\sum \nolimits_{j=1}^{i}x_j\right) = \tilde{p}^i, \; i=1,\;\dots,\;m,\\
& & & T = \sum \nolimits_{i=1}^m x_i,\\
& & & p \in \mathcal{P}_{T},
    \label{eqn:planning_general}
\end{aligned}
\end{equation}
where $\mathbf{x}=\left[\begin{array}{ccc}
x_1&\cdots&x_m
\end{array}\right]$ is the time allocation over the $m$ segments between waypoints, and the function space $\mathcal{P}_{T}$ is the set of all feasible trajectories, \ie, all trajectory functions that the quadrotor can successfully track over $[0,T]$.
By utilizing a piecewise polynomial function to describe the trajectory, waypoint constraints can be applied conveniently at the start and end of each segment, and differentiability constraints can be guaranteed by appropriately selecting the order of continuity between segments.
Using differential flatness of the idealized quadcopter dynamics, we know that for any feasible trajectory, $p_r \in \mathcal{C}^4$ and $p_\psi \in \mathcal{C}^2$ over the interval $[0,T]$~\cite{mellinger2011minimum}.
However, more constraints should be incorporated when considering fast and agile trajectories.
Critically, the complete system including aerodynamics, sensor and actuation hardware, and estimation and control software must be considered, instead of solely the idealized vehicle dynamics.
This is necessary because all aspects of the system affect tracking performance during demanding maneuvers, \eg, as follows:
\begin{itemize}
	\item Significant aerodynamic forces and momenta act on the vehicle and need to be accounted for in control inputs.
	In contrast, vehicle aerodynamics can typically be neglected in low-speed flight.
	\item State estimation errors due to delay and phase lag are exacerbated by the fast-changing vehicle state. Moreover, sensors measurements may incur additional noise due to large accelerations.	\item Actuation delay and bandwidth limitations, such as the mechanical time constant of the motors, inhibit tracking of fast-changing control inputs.
	\item Battery internal resistance causes voltage to drop under large currents. Consequently, very high motor speeds can only be maintained for limited consecutive time periods.
\end{itemize}
What results is a set of non-convex feasibility constraints with memory (i.e., depending on present and past states and control inputs).
These characteristics of $\mathcal{P}_T$ make the planning problem hard to solve, even when using nonlinear optimization techniques such as direct collocation or shooting methods.
Additionally, it is important to note that in practice no trajectory can be tracked perfectly due to stochastic sensing and actuation imperfections, so the definition of $\mathcal{P}_T$ must incorporate an error bound.

Popular methods for trajectory planning avoid the complicated feasibility constraints by reformulating the optimization problem such that 
feasibility is the objective instead of a constraint~\cite{mellinger2011minimum,richter2016polynomial}.
In practice, this is achieved by minimizing high-order derivatives of the trajectory function.
Particularly, minimizing the fourth-order derivative of position, \ie, snap, and the second-order derivative of yaw results in a minimization of the required angular accelerations.
This reduces trajectory agility and may thereby prevent activation of the aforementioned feasibility constraints.
The time to complete each trajectory segment is now treated as a constraint instead of the objective, leading to the following optimization problem:
\begin{equation}
\begin{aligned}
&\underset{p}{\text{minimize}}
& & \sigma(p,\sum \nolimits_{i=1}^{m}x_i)\\
&  \text{subject to} & & p(0) = \tilde{p}^0, \\
&   & & p\left(\sum \nolimits_{j=1}^{i}x_j\right) = \tilde{p}^i, \; i=1,\;\dots,\;m,
\label{eqn:minsnap}
\end{aligned}
\end{equation}
where 
\begin{equation}\label{eqn:smoothness}
\sigma(p,T)=\int_{0}^{T} \mu_r \norm{\frac{d^4 p_{r}}{dt^4}}^2 + \mu_\psi \Big(\frac{d^2 p_\psi}{dt^2}\Big)^2 dt
\end{equation}
with $\mu_r$ and $\mu_\psi$ weighing parameters.
This optimization problem can be formulated as an unconstrained quadratic program, and solved efficiently using matrix multiplications~\cite{richter2016polynomial}.
We define the function
\begin{equation}\label{eqn:min_snap_smoothness}
p = \chi(\mathbf{x},\tilde {\mathbf{ p}}),
\end{equation}
which --- provided a time allocation $\mathbf{x}$ and waypoints $\tilde{\mathbf{p}}$ --- gives a corresponding minimizer trajectory of \eqref{eqn:minsnap}.

Minimum-snap trajectory generation is a two-step process based on \eqref{eqn:minsnap}. First, the minimum-snap trajectory for a (very large) initial guess of the total trajectory time is found, as follows:
\begin{equation}
\begin{aligned}
&\underset{\mathbf{x}\in \realR^m_{\geq 0}}{\text{minimize}}
& & \sigma\left(\chi(\mathbf{x},\tilde {\mathbf{ p}}),T\right)\\
&  \text{subject to} & & T =\sum \nolimits_{i=1}^m x_i.
\label{eqn:minsnap-2}
\end{aligned}
\end{equation}
Next, the obtained time allocation is scaled down, \ie,
\begin{equation}
\begin{aligned}
&\underset{\eta\in\realR_{> 0}}{\text{minimize}}
& & T\\
&  \text{subject to} & & T =\sum \nolimits_{i=1}^m \eta x_i,\\
& & & \chi(\eta \mathbf{x},\tilde{\mathbf{p}}) \in \mathcal{P}_{T}.
\label{eqn:minsnap-3}
\end{aligned}
\end{equation}
The feasibility check is typically done based on control inputs obtained from differential flatness of the idealized quadrotor dynamics~\cite{mellinger2011minimum}.
While this method can generate fast, feasible trajectories, it does not attain time-optimality for several reasons.
Firstly, the time allocation ratio obtained in \eqref{eqn:minsnap-2} may not be the optimal as the total trajectory time is decreased, \ie, there may exist an alternative allocation ratio that will enable lower feasible total trajectory time.
Secondly, the method uses a simple feasibility model that fails to consider the trajectory as a whole, as is required for a realistic feasibility check.

Our proposed algorithm searches for a time-optimal trajectory by directly addressing the two major shortcomings of minimum-snap trajectory generation.
It builds a realistic, probabilistic feasibility constraint model that considers the trajectory as a whole, and uses this model to find the time-optimal trajectory segment time allocation.

\subsection{Bayesian Optimization}
Bayesian optimization, or \textit{BayesOpt}, is a class of algorithms that use machine learning techniques for solving optimization problems with objective or constraint functions that are expensive to evaluate.
It uses a Gaussian process (GP) model to build a \textit{surrogate model} that approximates these objective or constraint functions based on (noisy) measurements.
The GP model enables quantification of the uncertainty in the surrogate model~\cite{williams2006gaussian}.
This uncertainty is used in the \textit{acquisition function}, which selects each next evaluation point.
The acquisition function must be designed with consideration for both reducing uncertainty in the surrogate model, and finding the optimum of the objective function.
For instance, if only the objective function is modeled, the acquisition function can be based on the expected improvement of the objective~\cite{mockus1978application}, the expected entropy reduction of the distribution over the location of the solution~\cite{hernandez2014predictive, wang2017max}, or the upper bound of the optimum~\cite{srinivas2012information}.
If constraint functions are also modeled, the acquisition function must consider the uncertainty of both objective and constraint function models, \eg, by using the product of expected objective improvement and probability of constraint satisfaction~\cite{gardner2014bayesian, gelbart2015constrained}.
In our proposed algorithm, Bayesian optimization is applied to model the feasibility constraints of the time-optimal trajectory planning problem.
Related applications can be found in literature, where Bayesian optimization has been applied to model feasibility functions and dynamics boundaries for control synthesis in uncertain systems~\cite{berkenkamp2016safe, rai2019using}.

In multi-fidelity Bayesian optimization, function evaluations of different fidelities can be combined.
For instance, a rough simulation or an expert's opinion may serve as low-fidelity model, while a high-accuracy simulation or real-world experiment serves as high-fidelity model.
The key idea is that overall efficiency is improved by combining cheap low-fidelity evaluations and expensive high-fidelity evaluations.
The Bayes optimization surrogate model must be modified to be able to combine multi-fidelity evaluations.
This modification can be as simple as a linear transformation between different fidelities~\cite{kennedy2000predicting, le2014recursive},
or may include a more advanced nonlinear space-dependent transformation~\cite{perdikaris2017nonlinear,cutajar2019deep}.
In addition to the surrogate model, the acquisition function must also be modified to incorporate multi-fidelity evaluations.
Examples of such modifications for practically all common acquisition functions can be found in existing literature~\cite{huang2006sequential, takeno2019multi, kandasamy2016multi, costabal2019multi}.
Multi-fidelity Bayesian optimization enables a trade-off between accuracy and (computational) efficiency, and as such finds use in various fields including analog circuit design~\cite{zhang2019efficient}, wing design~\cite{rajnarayan2009trading}, and control synthesis~\cite{marco2017virtual, rai2019using}.

Although Bayesian optimization aims to contain the number of required function evaluations, it may still suffer from large computation cost as the number of data points increases.
This is mainly due to inversion of a covariance matrix including all data points for uncertainty quantification in the surrogate function, leading to a computational cost proportional to the number of data points cubed.
Particularly multi-fidelity Bayesian optimization may suffer from this issue, as the number of data points can quickly increase by adding low-fidelity evaluations.
The problem is addressed by the inducing points method, which uses a set of pseudo-data points for uncertainty quantification~\cite{snelson2006sparse, hensman2013gaussian}.
The pseudo-data points are selected to minimize the Kullback Leibler (KL) divergence between the function posterior given all data points (including the pseudo-data points), and the function posterior given only the pseudo-data points.
Since the number of pseudo-data points is much smaller than the number of actual data points, the inducing points method can greatly improve algorithm performance as the size of the data set increases.
The method was extended to the classification problem by \citet{hensman2015scalable}, and applied to multi-fidelity Bayesian optimization by \citet{cutajar2019deep}.
Its efficiency can be increased further by GPU acceleration~\cite{gardner2018gpytorch}.

\section{Algorithm} \label{sections:algorithm}

Our proposed algorithm uses Bayesian optimization to efficiently minimize the total trajectory time $T$ by approximating the feasibility constraints of $\mathcal{P}_T$ using multi-fidelity evaluations from various sources, such as simulation and real-world flight experiments.
Unlike typical minimum-snap trajectory planning, the algorithm maintains the ideal planning formulation of \eqref{eqn:planning_general} with minimum time as the objective and feasibility as a constraint.
We exploit the fact that any time allocation between given waypoints can be mapped to a smooth trajectory using \eqref{eqn:min_snap_smoothness}.
This enables us to transform the problem of finding the time-optimal trajectory to the problem of finding the optimal time allocation over segments.
Therefore, we can formulate the following optimization problem on the finite-dimensional space $\realR^m_{\geq 0}$ that is the set of all possible time allocations:
\begin{equation}
\begin{aligned}
&\underset{T, \; \mathbf{x}\in \realR^m_{\geq 0}}{\text{minimize}}
& & T\\
&  \text{subject to} & & \chi(\mathbf{x},\tilde {\mathbf{ p}})\in\mathcal{P}_T,\\
& & & T = \sum \nolimits_{i=1}^m x_i.
\label{eqn:alg_general_formulation}
\end{aligned}
\end{equation}
This formulation is based on the assumption that if there exists a feasible trajectory, \ie, a member of the set $\mathcal{P}_T$, that attains waypoints $\tilde{\mathbf{p}}$ with time allocation $\mathbf{x}$, then $\chi(\mathbf{x},\tilde {\mathbf{ p}})\in\mathcal{P}_T$.
This assumption is reasonable, since \eqref{eqn:minsnap} optimizes for feasibility (by minimizing snap),
so it is unlikely that its optimum is infeasible while there exists a feasible trajectory subject to the same time allocation and waypoints.



We use Gaussian process classification (GPC) to learn a surrogate model of the feasibility constraint in \eqref{eqn:alg_general_formulation}.
The surrogate model combines results from sequential experiments at $L$ fidelity levels denoted by $l \in \{l^1,l^2,\dots,l^L\}$, where $l^1$ is the fidelity level of the lowest-fidelity experiment and $l^L$ is the fidelity level of the highest-fidelity experiment.
An experiment $f_l(\mathbf{x})$ evaluates the feasibility of the trajectory $\chi(\mathbf{x},\tilde {\mathbf{ p}})$, resulting in a classification $y \in \{\operatorname{feasible}, \operatorname{infeasible}\}$.
Evaluations at fidelity level $l$ are gathered in dataset $\mathcal{D}_l = \left\{\left(\mathbf{x}_i, y_i\right)\right\}_{i=1,\dots,N_l}$, where $N_l$ is the number of evaluations at fidelity level $l$.
The GPC probability model at fidelity level $l$, \ie, $P_l(y=\operatorname{feasible}|\mathbf{x})$, is based on data points from fidelity levels $l$ and lower.
The overall objective is to find the optimal time allocation that satisfies the feasibility constraint at the highest fidelity level with sufficient confidence, \ie, $P_{l^L}(y=\operatorname{feasible}|\mathbf{x}^*) \geq h$.

Each evaluation is selected based on the acquisition function, which considers \textit{exploration} and \textit{exploitation} with the goal of maximizing the efficiency of the overall optimization process.
Based on the current trained surrogate model, it estimates the expected improvements in objective function value and model accuracy.
%
%
Combining these two aims and considering the cost of evaluations at each fidelity level, we formulate the following subproblem to obtain the data point and fidelity level of the next evaluation:
\begin{align}
    \mathbf{x}_{i}, l_{i} = \argmax_{\mathbf{x}\in\mathcal{X},l\in\{l^1,\dots,l^L\}} \alpha(\mathbf{x},l|\mathcal{D}),
    \label{eqn:alg_acquisition_general}
\end{align}
where $\alpha$ is the acquisition function, $\mathcal{D} = D_{l^1} \cup\cdots D_{l^L}$ is the data set of all past evaluations, and $\mathcal{X}$ is the set of candidate solutions.
By iteratively improving the feasibility model and minimizing the objective function, our algorithm searches for the time allocation that minimizes total trajectory time.
An overview of the complete algorithm is given in Algorithm \ref{alg:main}, and its major components are detailed in ensuing sections.
%
\begin{algorithm}
\SetAlgoNoLine
\DontPrintSemicolon
\LinesNumbered
\SetSideCommentRight
\KwIn{\begin{tabular}[t]{l}
        Multi-fidelity experiments $f_l$, $l=l^1,\dots,l^L$;\\
        acquisition function $\alpha$
    \end{tabular}
        }
Find initial solution by solving \eqref{eqn:minsnap-3}\;
Initialize multi-fidelity dataset $\mathcal{D}_l,\:\forall l\in\{l^1,\dots,l^L\}$\;
\For{$i=1,\cdots$}{
    Build the surrogate model $\mathcal{M}_l,\:\forall l\in\{l^1,\dots,l^L\}$\;
    Generate candidate solutions $\mathcal{X}$\;
    $\mathbf{x}_{i}, l_{i} \gets \argmax_{\mathbf{x}\in\mathcal{X},l\in\{l^1,\dots,l^L\}} \alpha(\mathbf{x},l|\mathcal{D})$\;
    $y_i \gets f_{l_i}(\mathbf{x}_i)$\; $\mathcal{D}_{l_{i}}\gets\mathcal{D}_{l_{i}}\cup\{(\mathbf{x}_{i}, y_{i})\}$\;
}
$T^* \gets \sum \nolimits_{i=1}^m x^*_i$\;
$p^* \gets \chi(\mathbf{x}^*,\tilde {\mathbf{ p}})$\;
\KwOut{$T^*$, $p^*$}
\caption{Multi-fidelity black-box optimization for time-optimal trajectory planning}
\label{alg:main}
\end{algorithm}
\setlength{\textfloatsep}{0pt}

\subsection{Multi-Fidelity Constraint Model}
Since GPC uses the nonparametric Gaussian process prior, it can flexibly approximate unknown constraints. 
Given data points $\mathbf{X}=\{\mathbf{x}_1,\cdots,\mathbf{x}_N\}$ with corresponding labels $\mathbf{y}=\{y_1,\cdots,y_N\}$ GPC predicts the label $y_*$ for a test point $\mathbf{x}_*$ in the form of a class probability $P(y_*|\mathbf{x}_*,\mathbf{X},\mathbf{y})$.
We define the GPC-based surrogate model $\mathcal{M}$ as the set of latent variables $\mathbf{f}=\begin{bmatrix}f_1,\cdots,f_N\end{bmatrix}$, the hyperparameters of covariance matrix $\theta$, and the hyperparameters of the inducing points $\mathbf{m}$ and $\mathbf{S}$, such that $\mathcal{M}=\left(\mathbf{f},\theta,\mathbf{m},\mathbf{S}\right)$.

To estimate the class probability, GPC maps the latent variables $\mathbf{f}$ onto the probability domain $[0,1]$ with a response function, such as $\Phi(x)=\int_{-\infty}^x \mathcal{N}(s|0,1)ds$.
The correlation between the latent variables $\mathbf{f}$ and the data points $\mathbf{X}$ is described by a Gaussian process prior assuming the distribution of $\mathbf{f}$ is given by $\mathcal{N}(0,K(\mathbf{X},\mathbf{X};\theta))$.
Each element of covariance $K(\mathbf{X},\mathbf{X})$ is nonparametrically estimated by the corresponding data points and a covariance kernel $K_{ij}(\mathbf{X},\mathbf{X})=k(\mathbf{x}_i,\mathbf{x}_j)$ with hyperparameter $\theta$.
Then, the distribution of the latent variables $\mathbf{f}$, or $P(\mathbf{f}|\mathbf{X},\mathbf{y})$, can be determined by maximizing the likelihood function
\begin{equation}\label{key}
\begin{aligned}
P(\mathbf{y},\mathbf{f}|\mathbf{X})&=\Pi_{i=n}^N P(y_n|f_n)P(\mathbf{f}|\mathbf{X})\\&=\Pi_{n=1}^N \mathcal{B}(y_n|\Phi(f_n))\mathcal{N}(\mathbf{f}|0,K(\mathbf{X},\mathbf{X})),
\end{aligned}
\end{equation}
where $\mathcal{B}(x)$ is the Bernoulli likelihood used to formulate $\Phi(f_n)$ as a probability distribution.
The hyperparameters of the covariance kernel are determined using the same loss function.

Once the distribution of $\mathbf{f}$ is obtained, one can use the same Gaussian process assumption to estimate the distribution of $f_*$, which is the latent variable corresponding to the class probability $y_*$.
The covariance of $\mathbf{X}$ and a test point $\mathbf{x}_*$ is modeled with the same covariance kernel, so that the distribution of the latent variable $f_*$ can be estimated as
\begin{gather}
    P(\mathbf{f},f_*|\mathbf{x}_*,\mathbf{X}) = \mathcal{N}\left(\begin{bmatrix}\mathbf{f}\\f_*\end{bmatrix}\left|0,
    \begingroup
    \setlength\arraycolsep{0.2pt}
    \begin{bmatrix}
        K(\mathbf{X},\mathbf{X}) & K(\mathbf{X},\mathbf{x}_*) \\
        K(\mathbf{x}_*,\mathbf{X}) & K(\mathbf{x}_*,\mathbf{x}_*)
    \end{bmatrix}
    \endgroup
    \right.\right),
    \label{eqn:gpc_1} \\
    P(f_*|\mathbf{x}_*,\mathbf{X},\mathbf{y}) = \int P(f_*|\mathbf{f},\mathbf{x}_*,\mathbf{X})P(\mathbf{f}|\mathbf{X},\mathbf{y})d\mathbf{f}, \label{eqn:gpc_2}
\end{gather}
where $P(f_*|\mathbf{f},\mathbf{x}_*,\mathbf{X})$ is obtained by marginalizing $\mathbf{f}$ in \eqref{eqn:gpc_1}. The resulting class probability is obtained by
\begin{gather}
P(y_*|\mathbf{x}_*,\mathbf{X},\mathbf{y}) = \int P(y_*|f_*)P(f_*|\mathbf{x}_*,\mathbf{X},\mathbf{y})d\mathbf{f_*}.
\label{eqn:gpc_3}
\end{gather}
For more details on GPC and its implementation, the reader is referred to \citet{nickisch2008approximations}.


We utilize the work by \citet{hensman2015scalable} to accelerate the optimization process using the inducing points method.
Estimating the distribution of the latent variables, \ie, $P(\mathbf{f}|\mathbf{X},\mathbf{y})$, requires the calculation of the inverse covariance matrix, which makes the computational complexity of the Gaussian process $\mathcal{O}(N^3)$, where $N$ is the number of data points.
By introducing an additional variational distribution $q(\mathbf{u})=\mathcal{N}(\mathbf{m},\mathbf{S})$, the inducing point method approximates $P(\mathbf{f}|\mathbf{X},\mathbf{y})\sim q(f):=\int p(f|\mathbf{u}) q(\mathbf{u}) d\mathbf{u}$.
The hyperparameters $\mathbf{m},\mathbf{S}$ represent the mean and covariance of the inducing points.
The method uses the following variational lower bound as loss function to determine the latent variables and the hyperparameters:
\begin{align}
\mathcal{L} = \sum_{n=1}^{N} \mathbb{E}_{q(f_n)} \left[ \log p(y_n|f_n) \right] 
- D_{KL} \left[ q(\mathbf{u}) \Vert p(\mathbf{u}) \right].
\label{eqn:alg_modeling_inducing_point}
\end{align}
In order to utilize the multi-fidelity optimization technique, we update the definition of the surrogate model and define the $l^j$-fidelity surrogate model ($j=1,\dots,L$) as
\begin{align}
    \mathcal{M}_{l^j} = \{f^{l^j}, [\theta^{l^1},\cdots,\theta^{l^j}], \mathbf{m}^{l^j}, \mathbf{S}^{l^j}\}.
    \label{eqn:alg_def_surrogate_model}
\end{align}
We use the multi-fidelity deep Gaussian process (MFDGP) by \citet{cutajar2019deep} as covariance kernel function to estimate the GP prior from multiple evaluation sources. 
The multi-fidelity Gaussian process has separate latent variables and inducing points $\mathbf{f}^{l^j},\:\mathbf{u}^{l^j}\sim\mathcal{N}(\mathbf{m}^{l^j},\mathbf{S}^{l^j})$ for each fidelity ${l^j}$.
Thus, we define the surrogate model separately for each fidelity model.
However, fidelity models share the hyperparameters of the covariance kernels.
Specifically, MFDGP models the relationship between adjacent fidelities with another Gaussian process as
\begin{equation}
    \begin{aligned}
        k_{l^j}(\mathbf{x},\mathbf{x}') = k_{{l^j},corr}(\mathbf{x},\mathbf{x}')(\sigma_{{l^j},linear} ^2 \:f_{l^{j-1}}(\mathbf{x})^T f_{l^{j-1}}(\mathbf{x}') \\
        + k_{l^j,prev}(f_{l^{j-1}}(\mathbf{x}), f_{l^{j-1}}(x'))) + k_{l^j,bias}(\mathbf{x},\mathbf{x}'),
    \end{aligned}
\label{eqn:alg_mfdgp_covariance}
\end{equation}
where $f_{l^{j-1}}$ is the Gaussian process estimation from the preceding fidelity level, $\sigma_{{l^j},linear}$ is a constant scaling the linear covariance kernel, and
$k_{{l^j},prev}$, $k_{{l^j},corr}$ and $k_{{l^j},bias}$ represent the covariance with the preceding fidelity, the space-dependent correlation function and the bias function, respectively.
Radial basis functions are used for these kernels. Their hyperparameters are trained by maximizing the following variational lower bound:
\begin{equation}
\begin{aligned}
    \mathcal{L} = \sum_{j=1}^{L} \Big(
    \sum_{n=1}^{N_{l^j}} \mathbb{E}_{q_{l^j}(f_n^{l^j})} \left[ \log p(y_n^{l^j}|f_n^{l^j}) \right] \\
    - D_{KL} \left[ q_{l^j}(\mathbf u^{l^j}) \Vert p( \mathbf u^{l^j}) \right] \Big),
\end{aligned}
\label{eqn:alg_modeling_mfdgp}
\end{equation}
where $N_{l^j}$ denotes the number of data points in the $l^j$-fidelity dataset $D_{l^j}$, and $q_{l^j}(f_n^{l^j}) := \int p(f_n^{l^j}|\mathbf{u}^{l^j})q(\mathbf{u}^{l^j})d\mathbf{u}^{l^j}$, which is the same formulation as used in \eqref{eqn:alg_modeling_inducing_point} for the hyperparameters of the $l^j$-fidelity model.
Note that inferencing of data points at $l^j$-fidelity requires estimation of the covariance kernel at $l^j$ and lower fidelities (cf. \eqref{eqn:alg_def_surrogate_model}), while the $l^j$-fidelity covariance kernel is trained based on the all data points at $l^j$ and higher fidelities.
%
We follow the implementation of the inducing point method by \citet{cutajar2019deep}, but utilize the likelihood approximation by \citet{hensman2015scalable} to extend the multi-fidelity kernel to the classification problem.
The overall GPC modeling is developed in the GPytorch framework with GPU acceleration~\cite{gardner2018gpytorch}.

\subsection{Acquisition Function}
We design the acquisition function \eqref{eqn:alg_acquisition_general} considering both \textit{exploration} to improve the surrogate model, and \textit{exploitation} to find the optimal solution.
In exploration, we aim to maximize the effectiveness of improving the surrogate model.
Inspired by \citet{costabal2019multi}, we select the most uncertain sample near the decision boundary of the classifier.
Since the latent function mean approaches zero at the decision boundary, this sample is found as the maximizer of
\begin{align}
\alpha_{explore}(\mathbf{x},l) = -\frac{|\mu_l(\mathbf{x})|}{\sigma_l(\mathbf{x})}C_l,
\label{eqn:alg_acquisition_explore}
\end{align}
where $C_l$ reflects the cost of an evaluation at fidelity level $l$.
We note that in practice the feasible set is connected --- feasibility of a time allocation implies feasibility of any time allocation that is element-wise larger --- so that exploration on the boundary suffices.
%
%
In exploitation, we utilize expected improvement with constraints (EIC) to quantify the expected effectiveness of a candidate data point based on the product of expected objective improvement and the probability of feasibility~\cite{gardner2014bayesian}.
We modify EIC for use in an experimental robotics application, where the cost of evaluation depends on the outcome, as attempting infeasible experiments may pose danger to the vehicle and its surroundings.
%
%
We consider not only the probability of success, but also the corresponding variance, as follows:
\begin{equation}
    \Tilde{P}_l(y=1|\mathbf{x}) = P_l (\mu_l(\mathbf{x}) - \beta\sigma_l(\mathbf{x}) \geq 0|\mathbf{x}),
    \label{eqn:alg_acquisition_exploit_lower_bound}
\end{equation}
where $\beta$ is a penalty on variance, and $\mu_l$ and $\sigma_l$ are respectively the mean and standard deviation of the posterior distribution $P(f|\mathbf{x}, \mathcal{D})$ obtained from \eqref{eqn:gpc_2}.
The resulting acquisition function is given by
\begin{equation}
    \alpha_{exploit}(\mathbf{x},l) = 
    \begin{cases}
        \alpha_{EI}(\mathbf{x}) \Tilde{P}_l(y=1|\mathbf{x}), &\text{if } \Tilde{P}_l(y=1|\mathbf{x}) \geq h_l\\
        0,              & \text{otherwise}\label{eqn:alg_acquisition_exploit}
    \end{cases}
\end{equation}
where $h_l$ is based on the cost of an infeasible evaluation.
Since the objective function is deterministic, so is the expected improvement, \ie, $\alpha_{EI}(\mathbf{x}) = \sum_i \Bar{x}_i - \sum_i x_i$ with $\bar{\mathbf{x}}$ the current best solution.

\begin{figure*}[!htb]
\centering
\begin{subfigure}[b]{0.28\textwidth}
    \includegraphics[width=\textwidth,trim=.0cm -1.0cm 0.6cm .0cm,clip]{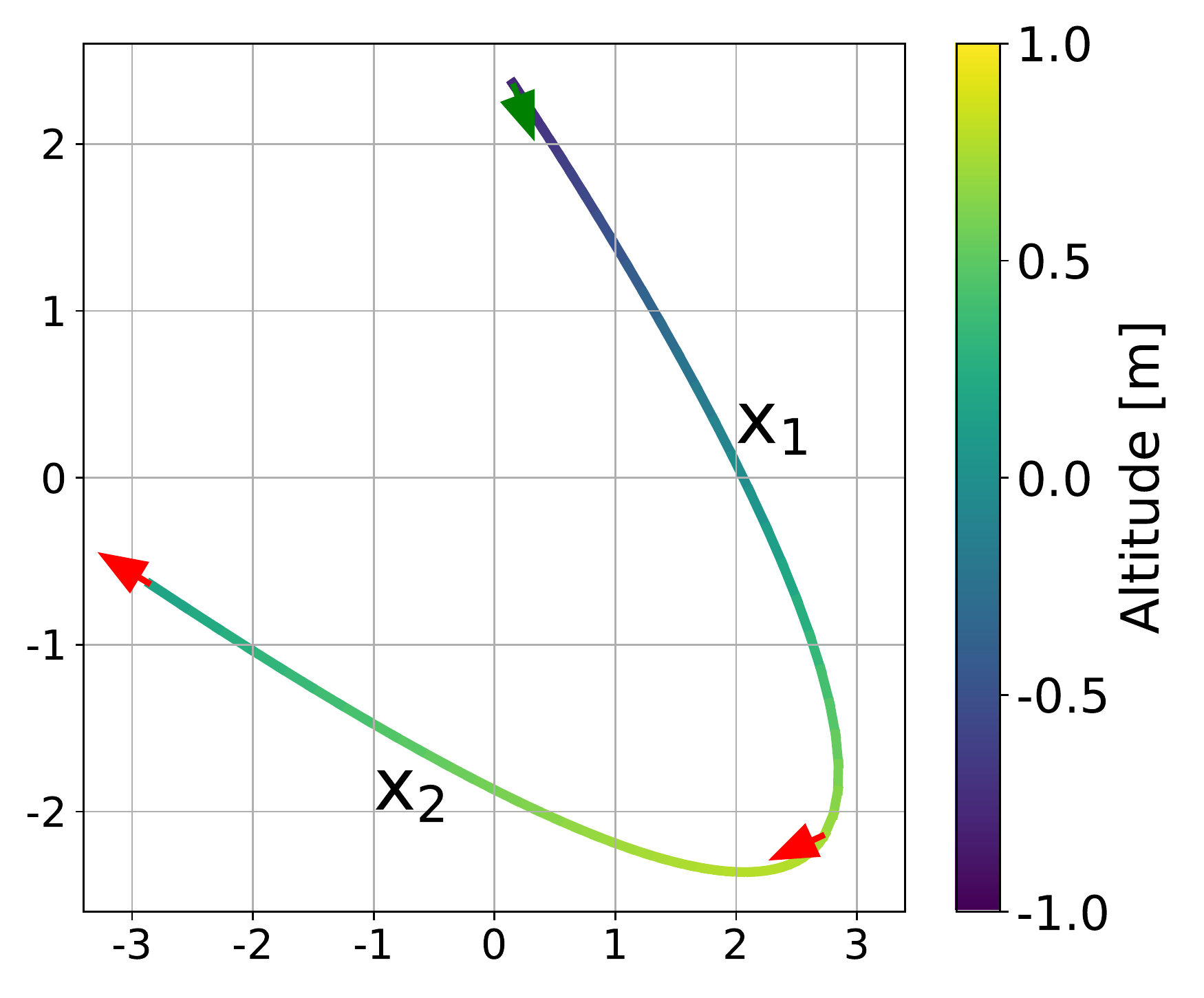}
    \caption{Trajectory with waypoints.}
    \label{fig:exp_two_segment_trajectory}
\end{subfigure}
\begin{subfigure}[b]{0.30\textwidth}
    \includegraphics[width=\textwidth,trim=.3cm .3cm 1.0cm .0cm,clip]{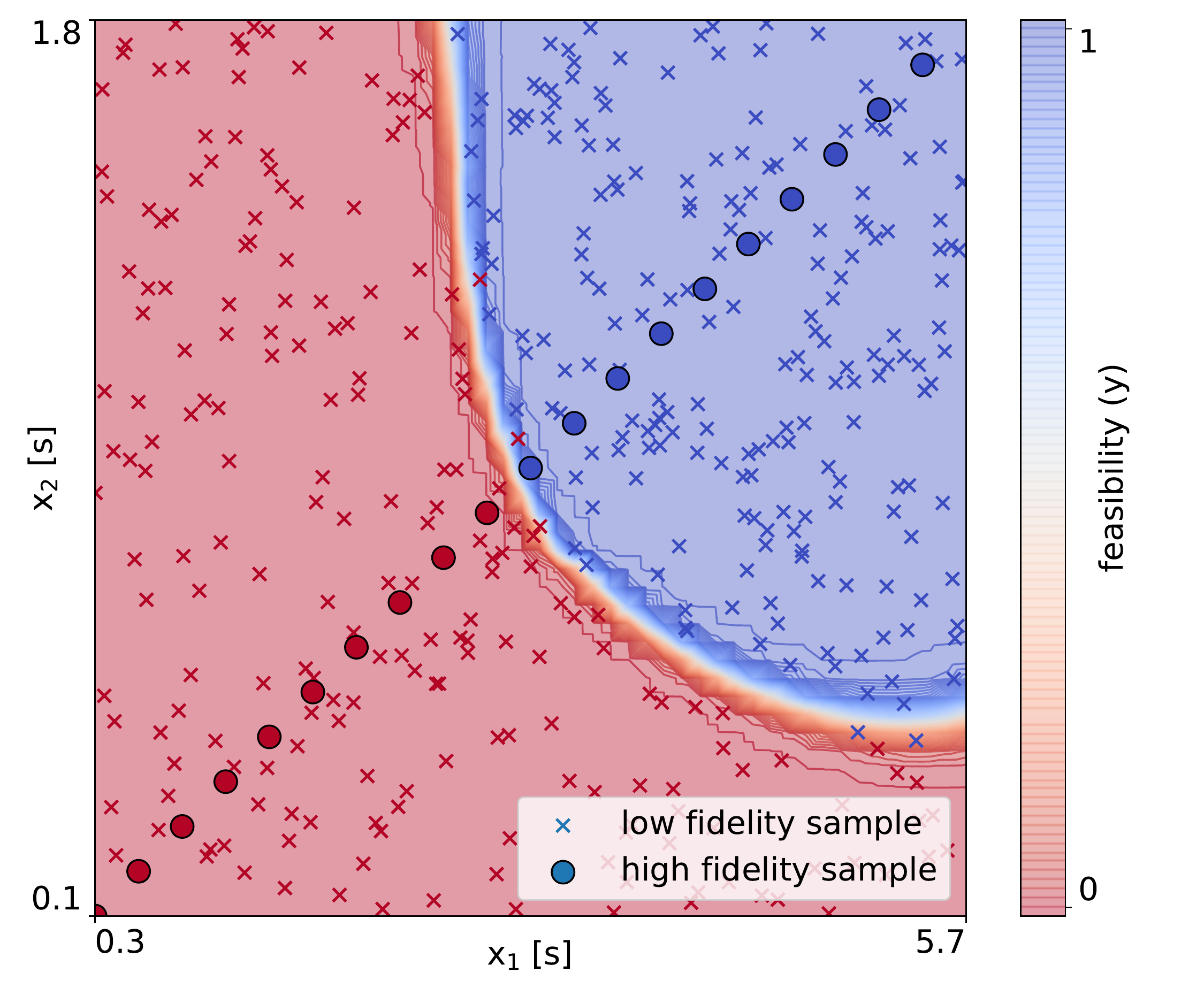}
    \captionsetup{width=0.9\textwidth, justification=centering}
    \caption{Initial medium-fidelity feasibility.}
    \label{fig:exp_two_segment_feasibility_init}
\end{subfigure}
\begin{subfigure}[b]{0.30\textwidth}
    \includegraphics[width=\textwidth,trim=.3cm .3cm 1.0cm .0cm,clip]{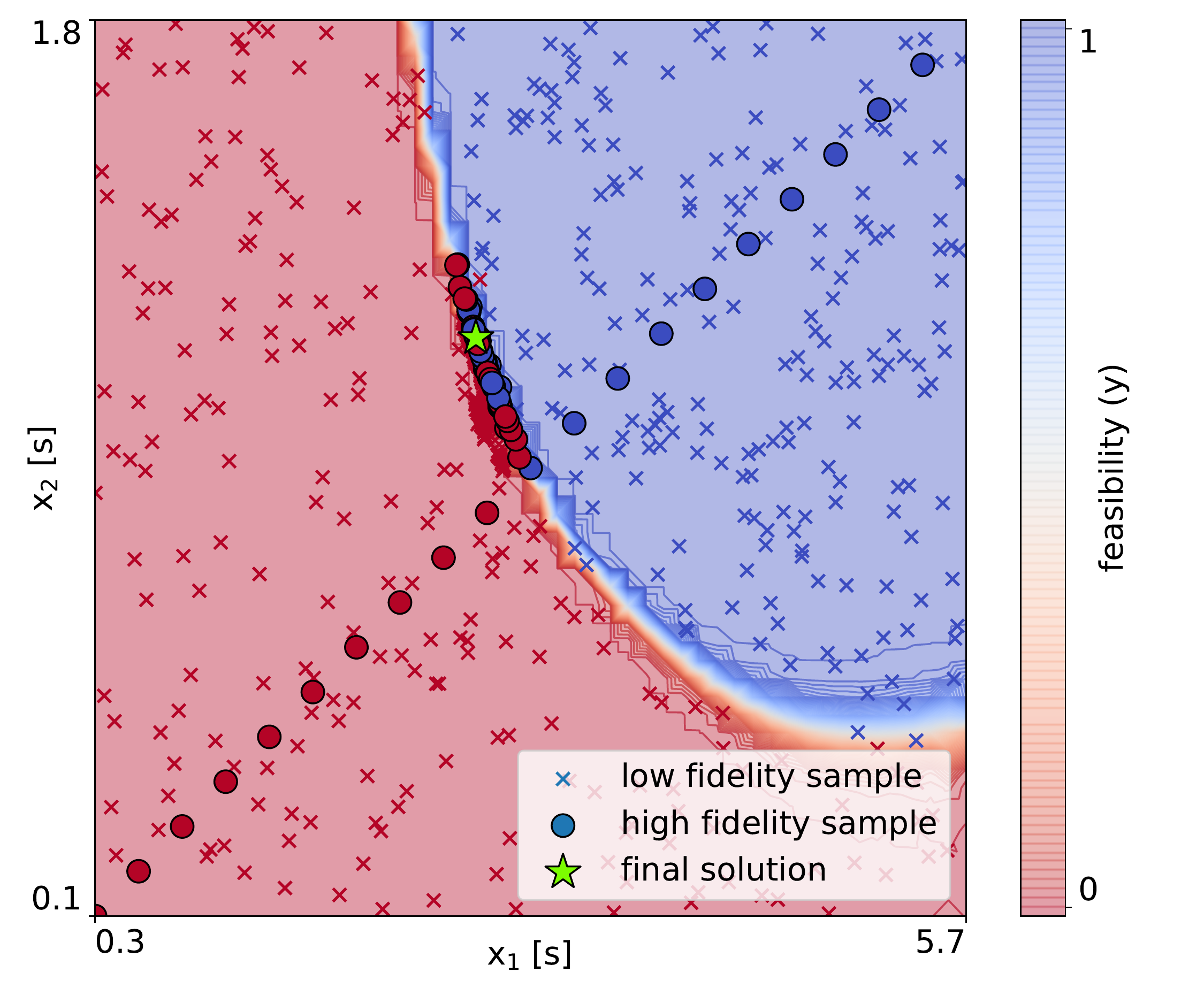}
    \captionsetup{width=0.9\textwidth, justification=centering}
    \caption{Final medium-fidelity feasibility.}
    \label{fig:exp_two_segment_feasibility_final}
\end{subfigure}
\caption{
    Waypoints and feasibility maps for two-segment trajectory optimization in the simulation environment.}
\label{fig:exp_two_segment_diagram}
\vspace{-5mm}
\end{figure*}

%
%
%
%
Finally, we combine \eqref{eqn:alg_acquisition_explore} and \eqref{eqn:alg_acquisition_exploit} to obtain
\begin{equation}
    \alpha(\mathbf{x},l) =  
    \begin{cases}
        \alpha_{exploit}(\mathbf{x},l), &\text{if } \exists x\in\mathcal{X} \:\text{s.t.}\:\alpha_{exploit}(x,l) > 0 \\
        \alpha_{explore}(\mathbf{x},l), &\text{otherwise}
    \end{cases}
    \label{eqn:alg_acquisition}
\end{equation}
Note that exploration only occurs if exploitation is ineffective. We found that this greedy approach works well in practice, as overly explorative searching leads to a large number of infeasible evaluations.

In case of a discontinuous acquisition function like \eqref{eqn:alg_acquisition}, Latin hypercube sampling (LHS) is often used to generate candidate solutions of \eqref{eqn:alg_acquisition_general}~\cite{srinivas2012information}.
However, we observe that this method may fail to generate an optimal solution because it does not consider the correlation between adjacent time allocations.
We address this by devising a sampling-based algorithm that uses a smooth perturbation of the current best solution.
For this purpose, we extend the notion of goal-convergent exploration by \citet{kalakrishnan2011stomp}.
This method generates smooth trajectories by minimizing the expected sum of acceleration at each collocation point.
%
%
%
%
Instead of acceleration, we minimize the expected jerk.
As time allocation is inversely related to flight speed, this approximately minimizes the jerk of the speed profile, and thereby the snap of the perturbation, which is favorable for maintaining feasibility. The perturbation vector $\mathbf{\epsilon}$ is found by solving the following semi-definite program:
\begin{equation}
\begin{aligned}
&\underset{\Sigma \in \realR^{m\times m}}{\text{minimize}}
& & \operatorname{trace}(A^TA \Sigma) \big(=\mathbb{E}[\mathbf{\epsilon} A^TA \mathbf{\epsilon}]\big)\\
&  \text{subject to} & & \mathbf{\epsilon} \sim \mathcal{N}(0,\Sigma),\\
& & & \Sigma \succeq 0, \;\Sigma_{ii}=\gamma,\label{eqn:alg_sampling_method}
\end{aligned}
\end{equation}
where $\gamma>0$ scales the variance of perturbations and $A$ is the third-order finite differencing matrix.
Finally, the set of $N_c$ candidate solutions is generated as $\mathcal{X} = \left\{\bar{\mathbf{x}}\odot(1+\epsilon_i)\right\}_{i=1,\dots,N_c}$ with $\epsilon_i \sim \mathcal{N}(0,\Sigma)$ and $\odot$ the element-wise product. Candidates with negative elements are rejected.

\subsection{Initialization}
Initial data points to build the surrogate model are generated around the trajectory found by \eqref{eqn:minsnap-3}, which may differ between fidelity levels, as feasibility constraints differ.
At fidelity levels with low evaluation cost, data point generation is done using LHS.
At fidelity levels where this method imposes prohibitive evaluation cost,
we use the fact that the initial trajectory is on the feasibility boundary.
As such, time allocations with the same ratio but shorter total time are infeasible, while time allocations with the same ratio but larger total time are feasible.
This enables generation of data points without any additional evaluation cost by scaling the initial trajectory.

The initial trajectory is also used for normalization of data points.
This is required because feasibility constraints differ between the different levels of fidelity, leading to bias in the dataset.
Moreover, time allocations between trajectory segments may be at different scales of magnitude, which may make the training process numerically unstable.
To resolve these issues, time allocation for each trajectory segment is scaled with the time allocation of the corresponding segment in the initial trajectory at the same fidelity level.

\subsection{Levels of Fidelity}
Evaluations at three different fidelity levels are used for quadrotor trajectory planning.
Low-fidelity evaluations are based on differential flatness of the quadrotor dynamics, which enables us to transform a trajectory and its time derivatives from the output space, \ie, position and yaw angle with derivatives, to the state and control input space, \ie, position, velocity, attitude, angular rate, and motor speeds~\cite{mellinger2011minimum}.
The resulting reference control input trajectory $u(t) = \zeta(p,t)$ would enable a hypothetical perfect quadcopter to track the trajectory $p$.
Feasibility of the control input values can be evaluated at relatively small computational cost, and as such can serve as a cheap, cursory evaluation of trajectory feasibility.
The set of feasible trajectories at fidelity level $l^1$ is defined as
\begin{equation}
\mathcal{P}^{l^1}_T = \: \Big\{p\Big|\zeta(p,t) \in \big[\underline u, \bar u\big]^4\;\;\;\;
\forall t \in \left[0,T\right]\Big\},
\label{eqn:feasibility_diff_flat}
\end{equation}
where $\underline u$ is the minimum and $\bar u$ is the maximum motor speed.

Medium-fidelity evaluations are obtained using the open-source multicopter dynamics and inertial measurement simulation by \citet{guerra2019flightgoggles} with the trajectory tracking controller by \citet{tal2018accurate}.
The feasible set is defined as
\begin{equation}
\begin{aligned}
\mathcal{P}^{l^2}_T = \Big\{p\Big|\norm{p_r(t)-r(t)}_2 \leq \bar r \land |{p_\psi(t)}-{\psi(t)}| \leq \bar \psi\;\; \\
\forall t \in \left[0,T\right]\Big\},
\end{aligned}
\label{eqn:feasibility_tracking_err}
\end{equation}
where $\bar r$ is the maximum allowable Euclidean position tracking error, and $\bar \psi$ is the maximum allowable yaw tracking error.
At this fidelity level, stochastic measurement and actuation noise,
motor dynamics, and
simplified aerodynamic effects are incorporated.
We note that these factors are typically unfavorable, leading to reduction of the feasible set compared to low fidelity.
At the same time, the controller may be able to perform adequate tracking even if reference control inputs are infeasible, so that neither feasible set is a subset of the other.
In order to account for stochastic effects, each evaluation consists of multiple simulations that all need to succeed for a trajectory to be deemed feasible.
%

High-fidelity evaluations are obtained from real-world experiments using a quadcopter and motion capture system.
At this fidelity level, each evaluation incorporates dynamics of the full system, including actuation and sensor systems, vehicle vibrations, unsteady aerodynamics, battery performance, and estimation and control algorithms.
This provides a highly accurate assessment of feasibility, but comes at great cost as it involves actual flying hardware and cannot be executed any faster than real-time.
We use the same controller as in simulation, and again perform multiple flights to account for stochastic effects.
The feasible set is identical to \eqref{eqn:feasibility_tracking_err}.
The overall objective of the algorithm is to find the time-optimal trajectory, subject to feasibility at the highest-fidelity model included.

\section{Experimental Results} \label{sections:experiment}
%
We evaluate the proposed algorithm in a \textit{simulation environment}, and in a \textit{hybrid environment} including simulation and real-world experiments.
In the simulation environment, we incorporate low-fidelity evaluations using reference control input feasibility, and medium-fidelity evaluations using the multicopter simulation.
In the hybrid environment, we incorporate medium-fidelity evaluations using the multicopter simulation, and high-fidelity evaluations using real-world flight experiments.
We chose not to incorporate low-fidelity evaluations, because their evaluation time is relatively very similar to medium-fidelity evaluations when compared to the time of flight experiments.
For all experiments, we set the maximum Euclidean position tracking error to 20 \si{cm} and the maximum yaw tracking error to 15 \si{deg}.
To reflect the difference in evaluation cost between evaluations at different fidelity levels, we set the parameters of the acquisition functions \eqref{eqn:alg_acquisition_explore} and \eqref{eqn:alg_acquisition_exploit} as $C_{l^j}=1$, $C_{l^{j+1}}=10$, $h_{l^j}=0.1$, $h_{l^{j+1}}=0.4$, and $\beta=3.0$ where $l^j$ is the lowest fidelity in each environment.
For the two-segment trajectory, yaw rate, velocity, acceleration, and jerk are constrained to zero at the first waypoint.
For each multi-segment trajectory, this is also the case at the final waypoint.

\begin{figure}
	\centering
	\includegraphics[width=0.48\textwidth]{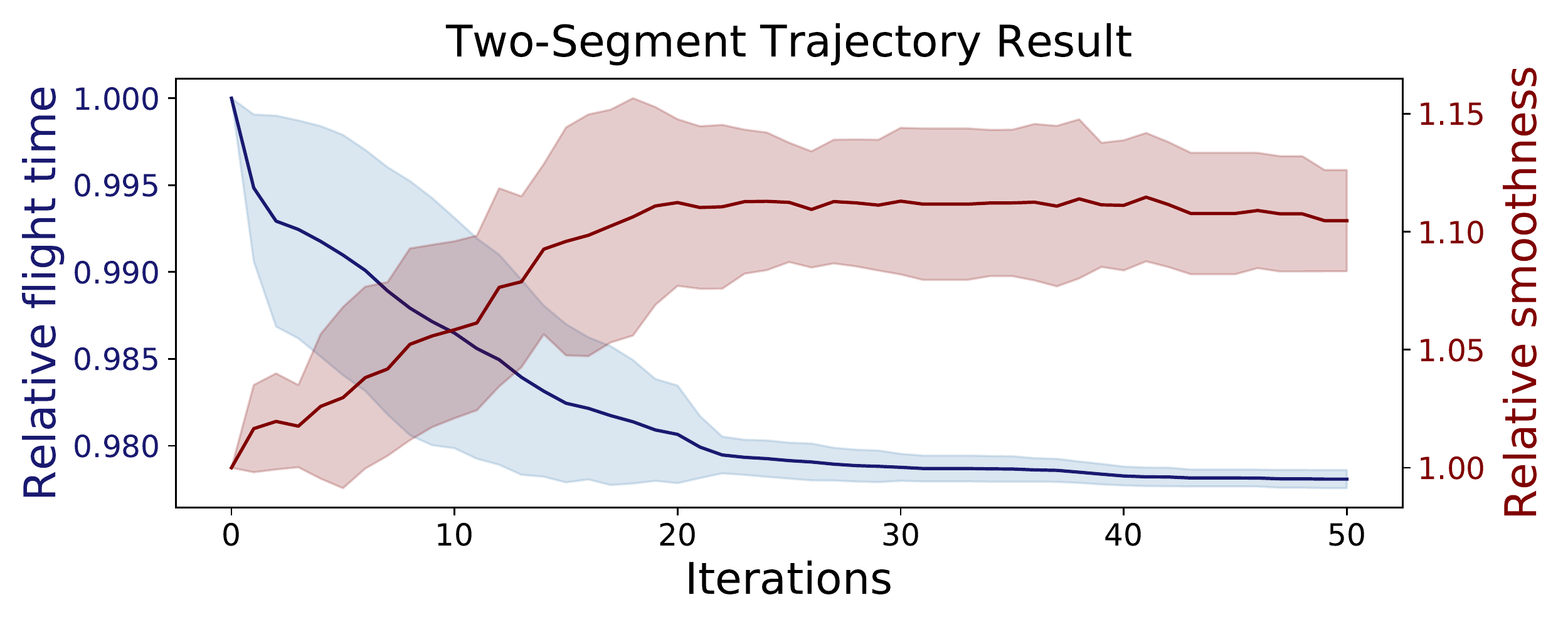}
	\caption{Mean and standard deviation of relative trajectory time and smoothness for the two-segment trajectory, obtained over 20 random seeds in the simulation environment.}
	\label{fig:exp_two_segment_result}
\end{figure}

To evaluate our algorithm, we compare the total trajectory time of the best solution at each iteration with the trajectory time obtained by \eqref{eqn:minsnap-3} using the same feasibility evaluation constraints.
Additionally, we compare the trajectory smoothness given by \eqref{eqn:smoothness}.
For this comparison, we scale the total trajectory time to the value obtained by \eqref{eqn:minsnap-3}, while maintaining the allocation ratio found by our algorithm.
Thereby, the comparison is based solely on the difference in allocation ratio, and not affected by total trajectory time.
In the corresponding plots, a higher value for smoothness indicates increased snap and yaw acceleration.


\subsection{Two-Segment Trajectory}
In order to illustrate the operation of our proposed algorithm, we first present results for the simple two-segment trajectory shown in Fig. \ref{fig:exp_two_segment_trajectory}.
These results were obtained in the simulation environment based on control input feasibility and the multicopter simulation.
The low-fidelity dataset is initialized using LHS of 400 data points, and the medium-fidelity dataset is initialized using 20 evenly scaled data points.

We consider each medium-fidelity evaluation along with preceding low-fidelity evaluations as a single iteration,
and limit the number of low-fidelity evaluations to 20 per iteration.
The limit is imposed to prevent the acquisition function from selecting too many successive low-fidelity evaluations, which may occur at iterations where the selected acquisition parameters do not behave well with the state of the medium-fidelity surrogate model.
Given the low dimension of the two-segment time allocation, we use LHS to generate candidate solutions for \eqref{eqn:alg_acquisition_general}.
We run the algorithm 20 times for 50 iterations using different random seeds.
Fig. \ref{fig:exp_two_segment_feasibility_init} and Fig. \ref{fig:exp_two_segment_feasibility_final} show respectively the initial and the final medium-fidelity feasibility maps for one of the runs.
Note that all medium-fidelity evaluations are in proximity of the final solution, as this promising region was first found by low-fidelity evaluations.
Fig. \ref{fig:exp_two_segment_result} improves the flight time by approximately 2\% compared to the initial trajectory obtained by \eqref{eqn:minsnap-3}.
The figure also shows that our algorithm selects time allocation ratios that result in increased snap and yaw acceleration at the same total trajectory time.
Despite this, the corresponding trajectories are still feasible at total trajectory times smaller than the initial trajectories.
This shows that our algorithm is indeed able to find an allocation ratio that compares favorably to the one found by \eqref{eqn:minsnap-2} and \eqref{eqn:minsnap-3}, and is thereby able to find feasible trajectories with smaller total trajectory time.
%
%
%
%
\begin{figure}
  \centering
  \includegraphics[width=0.48\textwidth,trim=.6cm .4cm .2cm .0cm,clip]{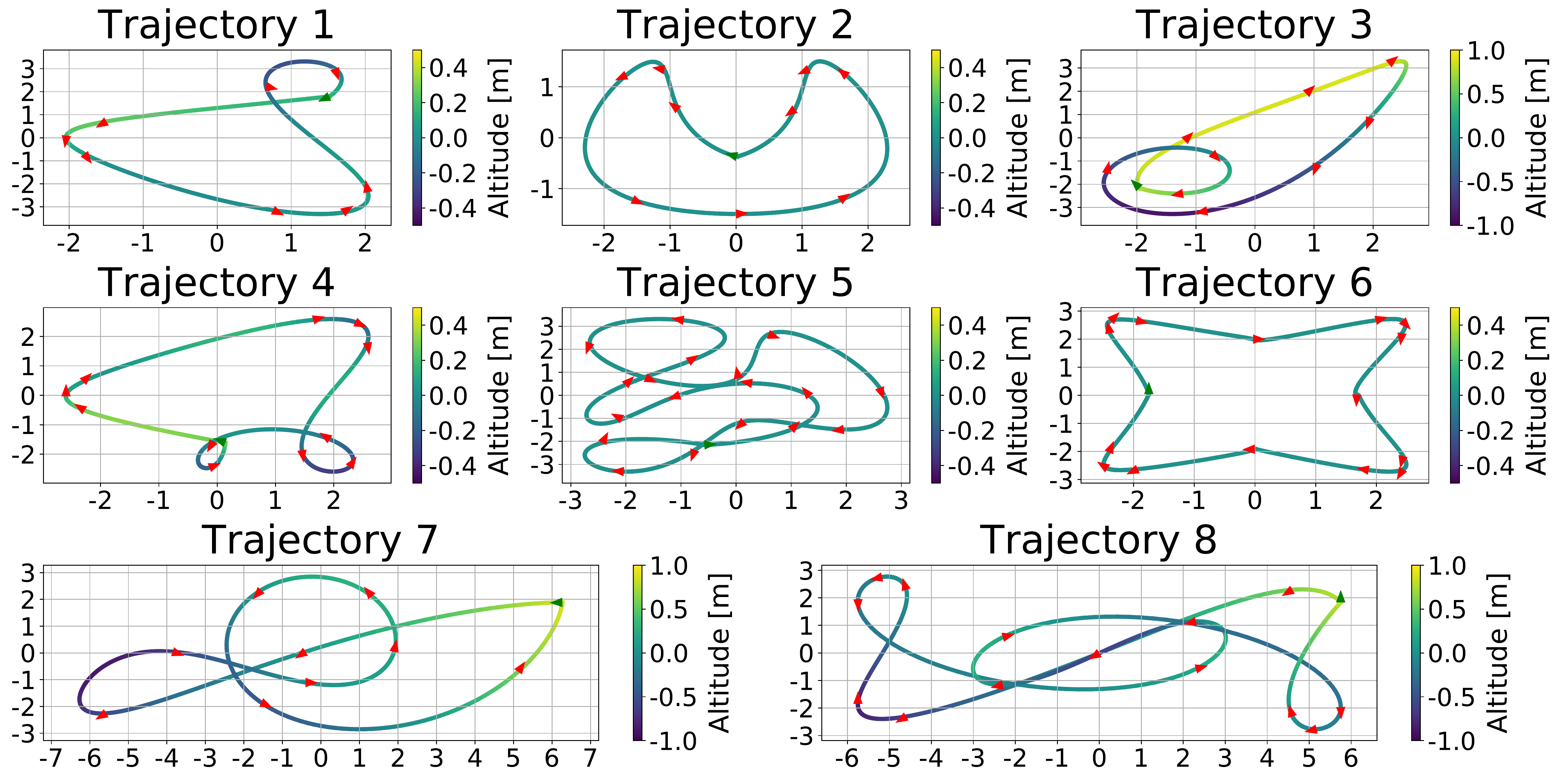}
  \captionsetup{width=0.45\textwidth}
  \caption{Multi-segment trajectories with starting point and subsequent waypoints indicated by green and red arrows, respectively.}
  \label{fig:exp_sample_traj}
\end{figure}
\subsection{Multi-Segment Trajectories}
Next, we apply our algorithm in the same simulation environment to the more complicated multi-segment trajectories shown in Fig. \ref{fig:exp_sample_traj}.
We increase the number of initial low-fidelity samples to 1000 and the maximum amount of low-fidelity evaluations per iteration to 50.
Given the increased dimensionality of the problem, we now use \eqref{eqn:alg_sampling_method} with $\gamma=0.2$ to generate the candidate solution set.

\begin{figure}
  \centering
  \includegraphics[width=0.48\textwidth]{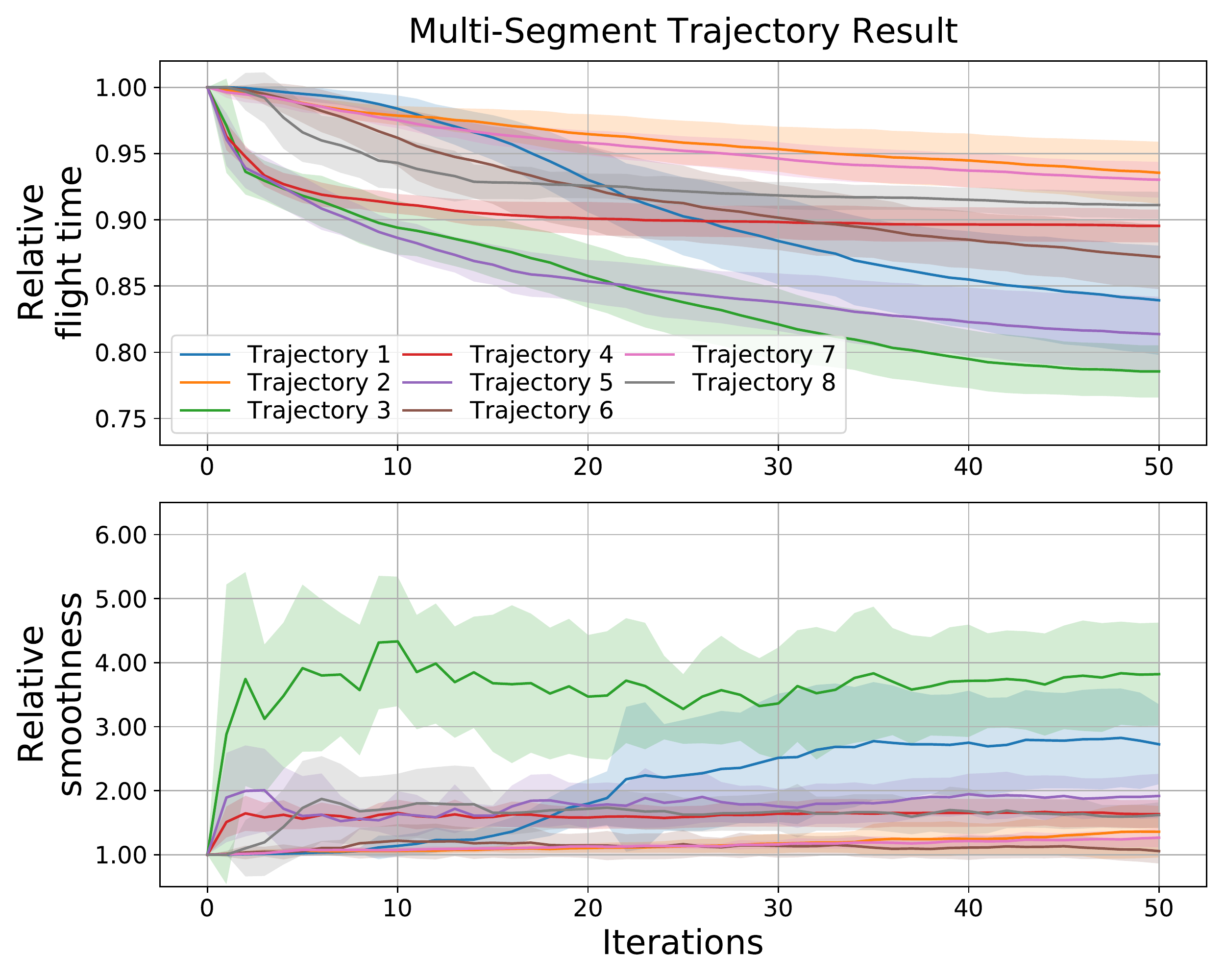}
  \caption{Mean and standard deviation of relative trajectory time and smoothness for the multi-segment trajectories, obtained over 20 random seeds in the simulation environment.}
  \label{fig:exp_multi_segment_result}
\end{figure}

The results in Fig. \ref{fig:exp_multi_segment_result} show that the algorithm is able to significantly reduce the trajectory time for each of the eight evaluated trajectories.
An average improvement of 22\% is achieved for \textit{Trajectory 3}.
The optimized trajectory is feasible despite having almost four times larger snap and yaw acceleration, which shows our algorithm's capability to find faster feasible trajectories compared to formulations that incorporate snap in the objective function.
%
%
%
%
%

\subsection{Real-World Flight Experiments}
We select three of the multi-segment trajectories shown in Fig. \ref{fig:exp_sample_traj} to evaluate in the hybrid environment with medium-fidelity evaluations using the multicopter simulation and high-fidelity evaluations using real-world quadrotor flight experiments.
The medium-fidelity dataset is initialized using LHS, and the high-fidelity dataset using the scaling method.

\begin{figure}[!t]
  \centering
  \includegraphics[width=0.48\textwidth]{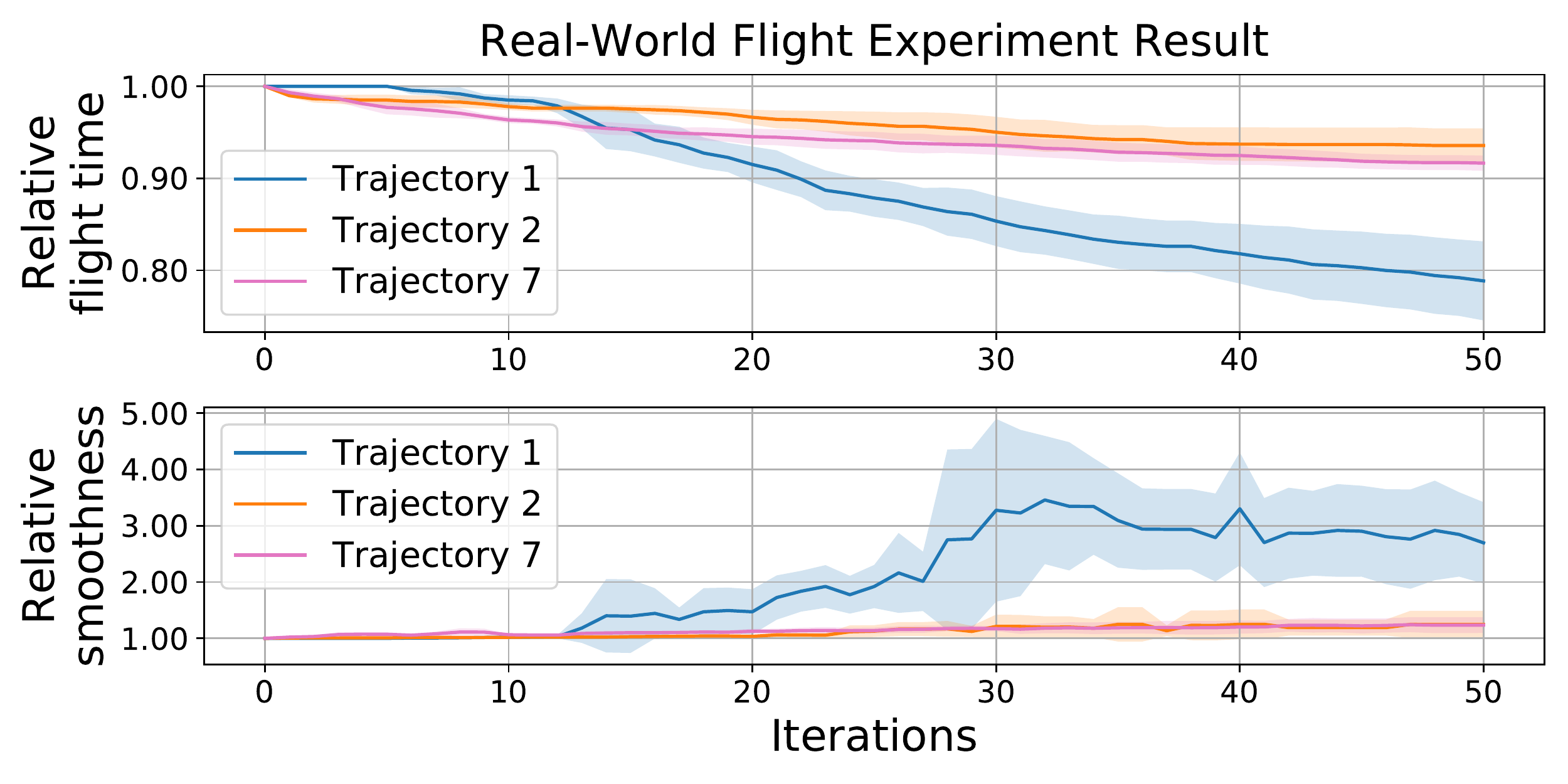}
  \caption{Mean and standard deviation of relative trajectory time and smoothness for the multi-segment trajectories, obtained over 5 random seeds in the hybrid environment using simulation and real-world flights.}
  \label{fig:exp_real_world_result}
\end{figure}

\begin{figure}[!t]
  \centering
  \includegraphics[width=0.48\textwidth]{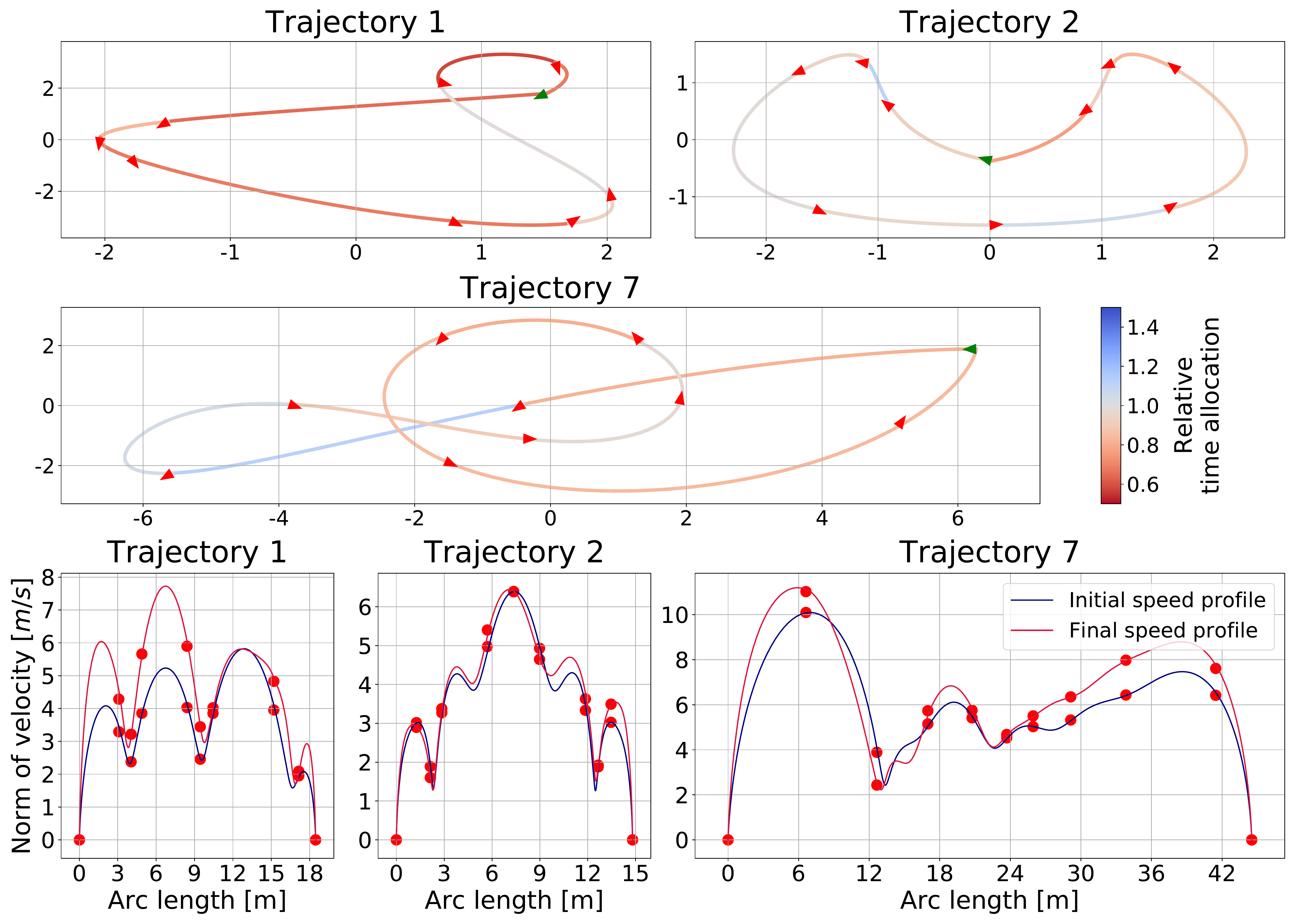}
  \caption{Relative time allocation and speed profiles of initial and optimized trajectories.}
  \label{fig:exp_real_world_result_alpha}
  \label{fig:exp_real_world_result_speed_profile}
\end{figure}

Comparison of Fig. \ref{fig:exp_multi_segment_result} and Fig. \ref{fig:exp_real_world_result} shows that results are consistent with those obtained in the simulation environment.
It can be seen that the algorithm achieves significant improvement using a limited number of real-world experiments.
Fig. \ref{fig:exp_real_world_result_alpha} shows the relative time allocation compared to the initial trajectory.
The time allocation is reduced for most segments, but also increased for some to maintain feasibility.
It can be seen that the vehicle decelerates before the turns and accelerates through them.
During experiments we observed that entering turns with reduced speed stabilizes tracking on the remainder of the trajectory, which can eventually reduce the overall flight time.
We also found that in general the vehicle is able to stabilize to static hover very quickly, allowing it to finish the trajectory quite aggressively.
Video of the flight experiments is available at \url{https://youtu.be/igwULi_H1Kg}.

\section{Conclusion} \label{sections:conclusion}
We presented our algorithm for modeling of quadrotor feasibility constraints and generation of time-optimal trajectories based on multi-fidelity Gaussian process classification.
The algorithm is able to incorporate evaluations from low and medium-fidelity sources such as analytical approximation and numerical simulation to minimize the number of required costly flight experiments.
Through extensive evaluation in simulation and real-life experiments, it was found that the algorithm is able to generate feasible trajectories that are significantly faster than those obtained from minimum-snap trajectory generation.
%
%
%





\bibliographystyle{plainnat}
\bibliography{references}

\end{document}